\documentclass[smallextended]{svjour3}       
\smartqed  
\usepackage{appendix}
\usepackage{amsmath}
\usepackage{graphicx}
\usepackage{array}
\usepackage{longtable}
\usepackage{natbib}
\usepackage{float}
\newcommand*\patchAmsMathEnvironmentForLineno[1]{%
\expandafter\let\csname old#1\expandafter\endcsname\csname #1\endcsname
\expandafter\let\csname oldend#1\expandafter\endcsname\csname end#1\endcsname
\renewenvironment{#1}%
{\linenomath\csname old#1\endcsname}%
{\csname oldend#1\endcsname\endlinenomath}}%
\newcommand*\patchBothAmsMathEnvironmentsForLineno[1]{%
\patchAmsMathEnvironmentForLineno{#1}%
\patchAmsMathEnvironmentForLineno{#1*}}%
\AtBeginDocument{%
\patchBothAmsMathEnvironmentsForLineno{equation}%
\patchBothAmsMathEnvironmentsForLineno{align}%
\patchBothAmsMathEnvironmentsForLineno{flalign}%
\patchBothAmsMathEnvironmentsForLineno{alignat}%
\patchBothAmsMathEnvironmentsForLineno{gather}%
\patchBothAmsMathEnvironmentsForLineno{multline}%
}
\usepackage[symbol]{footmisc}
\def\correspondingauthor{\footnote{Corresponding author.}}
\begin{document}

\title{The Unsupervised Method of Vessel Movement Trajectory Prediction
}


\author{Chih-Wei Chen         \and
        Charles Harrison \and Hsin-Hsiung Huang\correspondingauthor{}
}


\authorrunning{Hsin-Hsiung Huang} 

\institute{C.-W. Chen \at
              Department of Applied Mathematics \\
              National Sun Yat-sen University\\
              \email{BabbageTW@gmail.com, chencw@math.nsysu.edu.tw}           
           \and
           C. Harrison \at
           Department of Statistics and Data Science\\
          University of Central Florida \\
             \email{charleswharrison@knights.ucf.edu}
             \and
            *H.-H. Huang \at
           Department of Statistics and Data Science\\
          University of Central Florida \\
             \email{Hsin.Huang@ucf.edu}
}

\date{Received: January 2020 / Accepted: }

\maketitle

\begin{abstract}
In real-world application scenarios, it is crucial for marine navigators and security analysts to predict vessel movement trajectories at sea based on the Automated Identification System (AIS) data in a given time span. 
This article presents an unsupervised method of ship movement trajectory prediction which represents the data in a three-dimensional space which consists of time difference between points, the scaled error distance between the tested and its predicted forward and backward locations, and the space-time angle. The representation feature space reduces the search scope for the next point to a collection of candidates which fit the local path prediction well, and therefore improve the accuracy. 
Unlike most statistical learning or deep learning methods, the proposed clustering-based trajectory reconstruction method does not require computationally expensive model training. 
This makes real-time reliable and accurate prediction feasible without using a training set. Our results show that the most prediction trajectories accurately consist of the true vessel paths. 
\end{abstract}

\section{Introduction}
The National Geospatial-intelligence Agency (NGA), in collaboration with the National Science Foundation, designed a set of challenge
problems that are based on the automatic identification system (AIS) maritime vessel data \citep{AIS}. 
The AIS is a collaborative self-reporting system that
ships over 300 tonnes and all passenger ships must have
installed on board, as mandated by the Safety of Life at
Sea convention issued by the International Maritime Organization (IMO).
The AIS data contain time-stamped information about a maritime
vessel’s movement, including latitude, longitude, course over ground, and speed over ground.
The AIS data were chosen for algorithm evaluation due to its expansive, spatio-temporal nature, and the fact that the data are systematically archived. The data clearly identify the movement of each vessel, through Maritime Mobile Service Identity (MMSI). This is not the case
for typical data collections in a threat environment, where although movements of multiple objects can be tracked, identities of these objects are not always known. Thus, the AIS data are a good analog for studying threat detection.
The data are in the form of time-sequenced nodes, where each node contains the timestamp, coordinates (latitude and longitude), speed, and direction of a vessel. The MMSI number is withheld, and the awardees are asked to develop algorithms that associate each node with a track, with a goal that associated tracks will duplicate true tracks. However, to facilitate
the awardees to become familiar with the AIS data and develop algorithms, training data with the an anonymized MMSI number (or Vessel Identifier, VID) is initially provided.
There is no pre-ordained approach that the awardees should take in developing their
algorithms for track association. We provided a sample track association algorithm, to
demonstrate one way of solving the challenge problems. The awardees have complete flexibility
in their approach to the challenge problems. The only requirement is that the results be prepared
in a specified format, to facilitate subsequent performance evaluation.
To conduct a comprehensive performance evaluation, we considered metrics that account for
(1) counts of erroneous tracks, (2) the continuity score, and (3) the completeness score.
Algorithms will be anonymized in evaluation.

The challenge problems will be administered in a deliberate manner. We plan to initially
distribute this problem-definition document, training data (with the VID, two simpler challenge
problem sets (without the VID)), and the sample track association algorithm to the awardees. As
mentioned above, the training data are provided, so that the awardees can become familiar with
the AIS data and develop algorithms. The two initial challenge problem sets are easier, in that the
number of true tracks is made known or the number of tracks is relatively low. Subsequent
challenge problem sets are more difficult, in that the number of tracks is relatively high or some
data gaps are present. The performance of all algorithms, after anonymization, will be
systematically evaluated and summarized using the proposed metrics. virus. 

The AIS data consist of messages  including (1) the identifier (VID) number, (2) time stamp, (3) latitude, (4) longitude, (5) course over ground (i.e., a vessel’s direction with respect to the surface of the earth),  and (6) speed over ground (i.e., a vessel’s speed with respective to the surface of the earth).
All time-sequenced nodes for the same vessel collectively define the track of that vessel. We evaluated performances of the proposed algorithm and other methods using the AIS data without the VID to associate each node with a vessel’s track based on time stamp, latitude, longitude, 
course over ground, and speed over ground. Only the proposed method provides desired prediction without acquiring a training model. 


\section{Challenges and related works}

The prediction of vessel trajectories using AIS data is 
challenging due to the following reasons. First, the sample size varies a lot from vessel to vessel. Second, the AIS data have varying noise patterns and irregular time-sampling. Both are very common in AIS.
According to the International Maritime Organization (IMO), the  state-of-the-art supervised
machine learning models including deep learning methods could not solve these issues.

In this paper, we addressed these issues and proposed an algorithm
that extracts and characterizes local information in AIS data streams.
More specifically, our main contributions are three-fold:
(1) The design of a novel big-data-compliant unsupervised
algorithm which automatically learns and extracts useful information from noisy and partial AIS data streams
on a regional scale;
(2) The joint exploitation of this architecture as a basis
for specific tasks using mathematical modeling which reconstructs and forecast trajectories;
(3) The demonstration of the proposed approach’s relevance
on real regional nearby Norfolk, Virginia and simulated data. We used AIS data collected by a global network of coastal AIS receivers. The first AIS dataset was collected from 14:00:00 (2:00:00 pm) to 17:59:58 (5:59:58 pm) in an area spanning from $36.906505^\circ$ to $37.049995^\circ$ in latitude and from $-76.329934^\circ$ to $-75.98009^\circ$ in longitude; the second AIS dataset has data collected from 14:00:00 (2:00:00 pm) to 17:59:59 (5:59:59 pm) in an area spanning from $36.906063^\circ$ to $37.049933^\circ$ in latitude and $-76.329982^\circ$ to $-75.98^\circ$ in longitude; the the third dataset was collected from 14:00:00 (2:00:00 pm) to 17:59:58 (5:59:58 pm) in an area spanning $36.906038^\circ$ to $37.04974^\circ$ in latitude, $-76.329979^\circ$ to $-75.980184^\circ$ in longitude.


There are some related works in the field of vessel trajectory prediction based on AIS data, especially regarding
trajectory reconstruction and forecasting and anomaly detection.
In this paper, the term trajectory reconstruction means both  reconstructing and forecasting trajectories. Early efforts for trajectory reconstruction includes linear interpolation, curvilinear interpolation \citep{Best} and its improvements \citep{Perera2012MaritimeTM,Schubert2008ComparisonAE}, and Recurrent
Neural Networks (RNNs)  \citep{Nguyen2018AMD}. They rely on a physical model of the movement
information such as speeds, directions, and time. They typically use the Speed Over Ground (SOG) and the Course Over Ground (COG). More
sophisticated methods suppose that vessel trajectories follow a
distribution and learn it from historical data \citep{Millefiori2016ModelingVK,Fusion2014}. Currently,
the state-of-the-art methods for trajectory reconstruction \citep{Mazzarella2015KnowledgebasedVP,Hexeberg2017AISbasedVT,Coscia2018MultipleOP} use the following typical three-step approach: i) the first
step involves a clustering method \citep{Lee:2007:TCP:1247480.1247546,journals/entropy/PallottaVB13} to cluster historical motion data into route
patterns, ii) the second step assigns the vessel to be processed
to one of these clusters iii) the third step interpolates or predicts
the vessel trajectory based on the route pattern of the assigned
cluster. These methods are suitable for the AIS data with long time and distances for training normal patterns and detecting velocity changes, whereas our data consists of short-term and distances trajectories which are difficult to be detected or identified from a trained stochastic process based models.





\section{Method: Next-Point Connection}

We first transform the longitude and latitude as the Universal Transverse Mercator (UTM) coordinates. Then we predict each location's label (VID) by the following proposed methods.

The next-point connection (NPC) classification algorithms use the distance defined
as
$$
\min_{i\in K}
d(\max_{s\leq t_0}E_i(s), O(t)),
$$
where $E$ is the estimated location and $O$ is the observed test location at time $t$, $i$ is the index of the nearest estimated points from the training points $K$ of each label, and $s$ is the set of variables that find the closest training points.
The algorithm of the proposed classification method:
\begin{itemize}
    \item Step 1. Find the closest location for each tested point's location from each label before the test point's time.

    \item Step 2. Estimate the selected points' next location given the their speed, direction, and the time difference between the selected training point and the test point.

    \item Step 3. Predict the label of the test point by its closest estimated point at Step 2.
\end{itemize}

We now derive an unsupervised clustering algorithm from the above classification algorithm. The next-point connection (NPC) clustering algorithms uses the distance defined
as
$$
\min_{i\in I}
d(\min_{s}d(E_i(s), O(t))),
$$
where $E$ is the estimated location and $O$ is the observed test location at time $t$, $i$ is the index of the nearest estimated points from the selected nearest points $I$ (we chose the closest 3 points), and $s$ is the set of variables that find the closest points.
As a result, we have a clustering method without using labels from a training set.
The algorithm of the proposed clustering method:
\begin{itemize}
    \item Step 1. Find the $k$ nearest points ($k=3$ in our data analysis) for each point according to the Euclidean distance with all the features.

       \item Step 2. Calculate the average speed and direction (course) of every test point and each of its $k$ nearest neighbors, and compute the time difference between every test point and each of its $k$ nearest neighbors. 
       
       \item Step 3. Compute the estimated location by the speed and direction in Step 2. Group the test point with the point closest to the test point's estimated location.
\end{itemize}

\subsection{Clustering-based Trajectory Reconstruction}

In order to reconstruct each vessel's trajectory, we propose a  clustering-based trajectory reconstruction (CBTR) algorithm based on the NPC clustering method as follows. 

Given AIS points $x_i$ at time $t_i$, with position $p_i=({\rm latitude}_i,{\rm longitude}_i)$, speed $v_i$ and course $c_i$.
For every $x_i$, we use the information of speed and course to choose the best possible next point (BPNP) $x_j$. If we cannot find a nearby next point of $x_i$, then we treat $x_i$ as an end point of a trail. We classify the trails by these end points. 


The algorithm of CBTR:
\begin{itemize}
    \item Step 1. (Set up candidates for BPNP.)  For any $x_i$, collect all points $x_j$ with $t_j$ satisfying $t_i+1\leq t_j\leq t_i+1000$; they are candidates for the best possible next point of $x_i$. Here the time parameter $t$ is measured in seconds.\\

     Remark: If the upper bound $1000$ is replaced by a small number, than some trajectories broken for long time periods will be treated as different vessels. Removing the upper bound does not affect our essential result, but the computation time is much longer. \\

    \item Step 2. (Find the BPNP from candidates.) 
    Compute the predicted next (forward) position $P^+(x_i)$ of $x_i$ by physical information. We use the velocity of $x_i$ at time $t_i$ to estimate the future position $P^+(x_i)$ of $x_i$. We define an error distance between $x_i$ and each of its candidate $x_j$. Later we will choose the $x_j$ with smallest error distance to be the BPNP of $x_i$. The pairs $(x_i,x_j)$ is called {\it moving} if the sum of speeds of $x_i$ and $x_j$ is larger than $3$ knots; otherwise the pair is called {\it steady}.
    
    {\it Case 1 for moving pairs}.

    For pairs of moving points, we first
    define a forward distance between $P^+(x_i)$ and $x_j$ based on their spatial and temporal differences (diff.) by
    $$d^+=(2\cdot10^{-6}\cdot\mbox{time diff.})^2+(\alpha\cdot\mbox{latitude diff.})^2+(\mbox{longitude diff.})^2.$$ Here $\alpha=69/(69.172\cdot\cos(\mbox{the average of latitudes})))$ is determined by UTM and $2\cdot10^{-6}$ is chosen based on experimentation.

\begin{figure}[H]
    \centering
    \includegraphics[scale = 0.9]{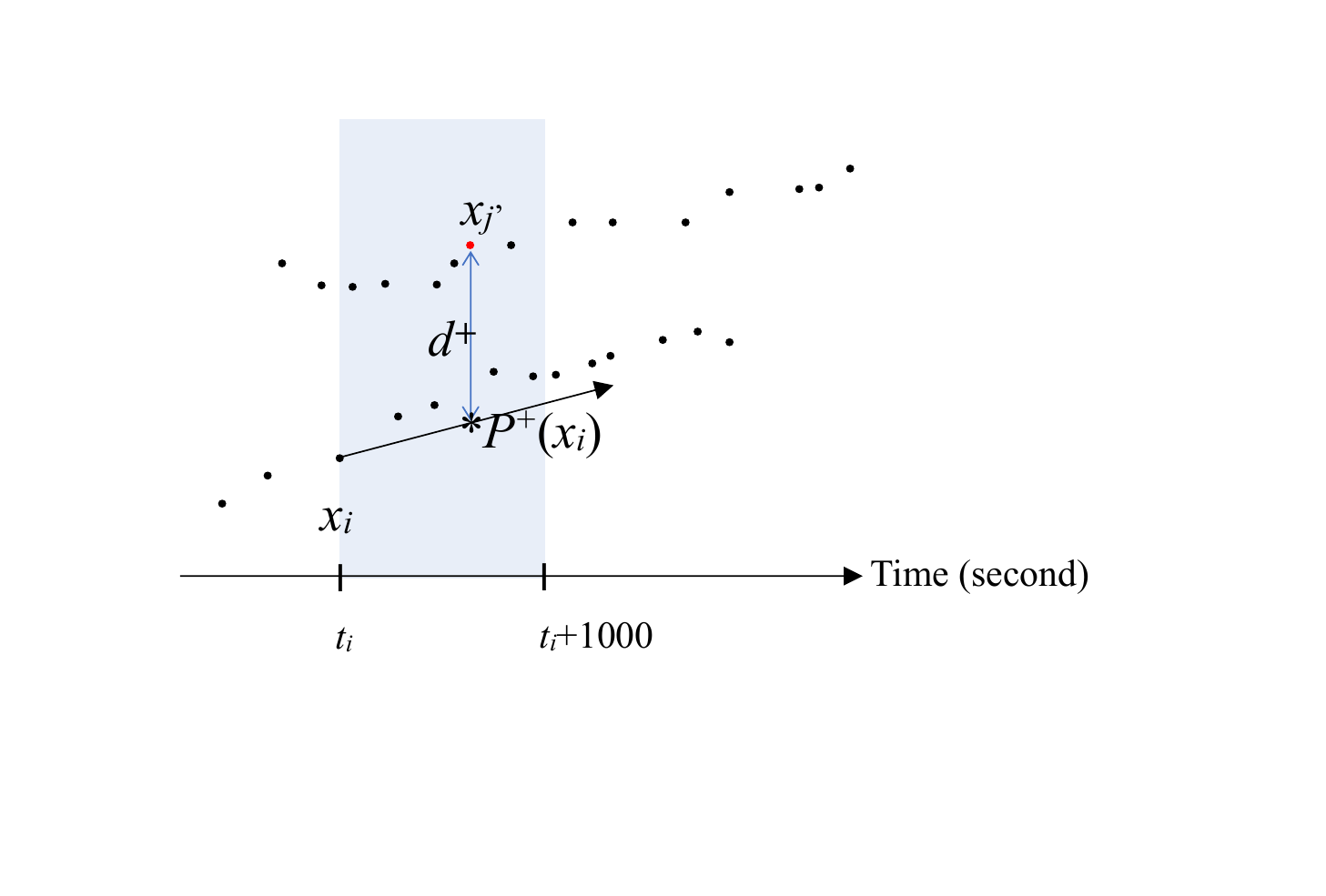}\vspace{-2.3cm}
    \caption{All candidates $x_j$ distribute in the shadow region. Use the velocity of $x_i$ to predict its future positions (on the arrow). The distance $d^+$ between $x_i$ and $P^+(x_i)$ given $x_j$. $P^+(x_i)$ is computed for each $t_i+1\leq t_j\leq t_i+10000$ seconds.}
\end{figure}

    Similarly, we define a backward distance $d^-$ to measure the difference between the predicted previous (backward) position $P^-(x_j)$ of each candidate and $x_i$. We expect that false candidates have large $d^-$ and can be eliminated later.\vspace{-2cm}
    
    \begin{figure}[H]
    \centering
    \includegraphics[scale = 1]{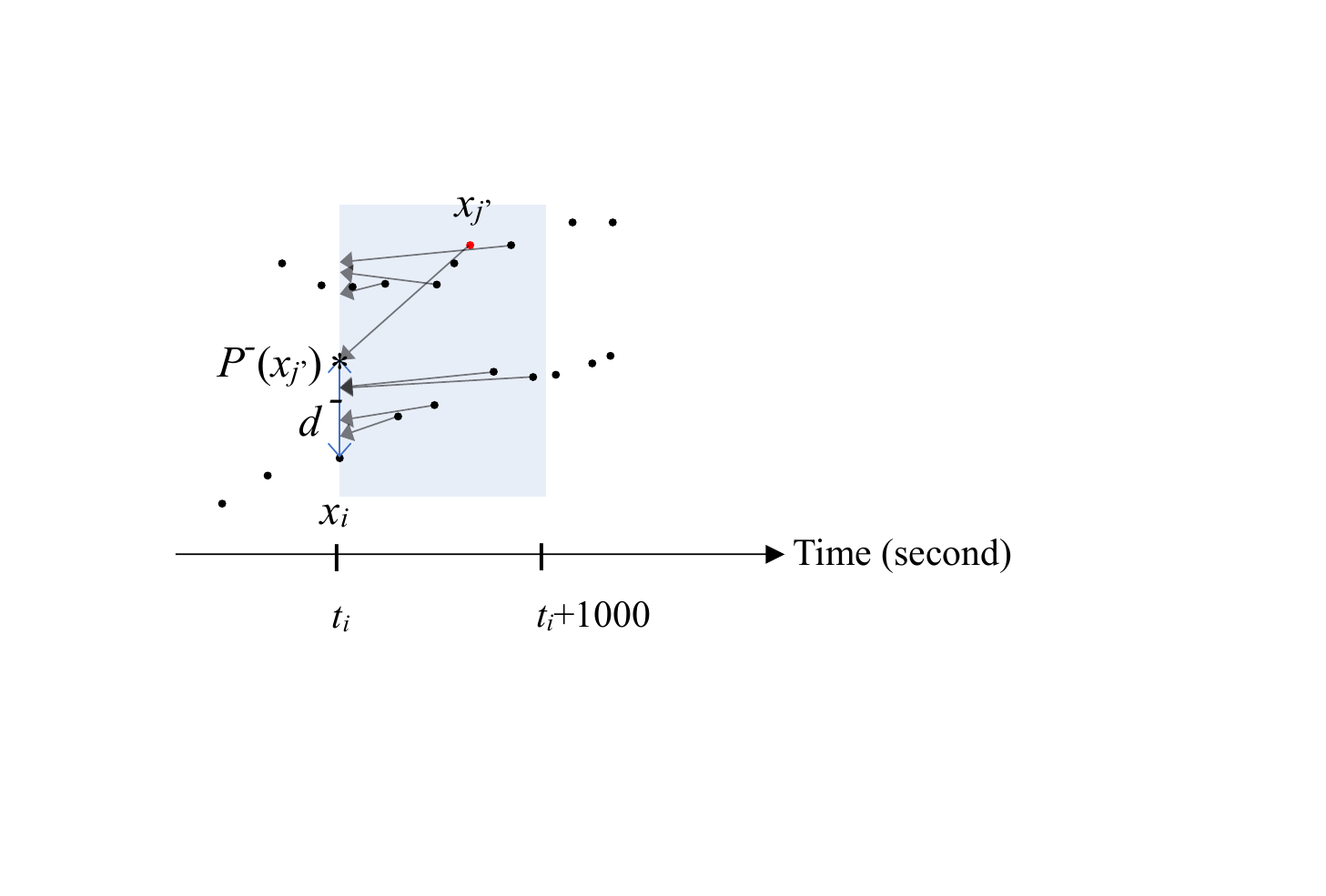}
    \vspace{-3cm}
    \caption{Use the velocity of each $x_j$ to estimate its previous position $P^-(x_j)$ at time $t_i$.}
\end{figure}

    On the other hand, we compute the space-time angle $\theta_{ij}$ between ${\overrightarrow{x_iP^+(x_i)}}$ and ${\overrightarrow{x_ix_j}}$, where the space-time vector is defined by 
    $$\left(10^{-5}\cdot\mbox{time diff.}, \alpha\cdot\mbox{latitude diff.}, \mbox{longitude diff.}\right)$$ with the tuning parameter $\alpha$.
    We kick off candidates with $\cos\theta_{ij}\leq 0.1$. Then we define the error distance $d_{ij}=\frac{1}{2}(d^++d^-)$ and choose the candidate $x_j$ with smallest $d_{ij}$ to be the BPNP of $x_i$. \vspace{-2cm}

    \begin{figure}[H]
    \phantom{A}\hspace{-1.5cm}\includegraphics[scale = 1]{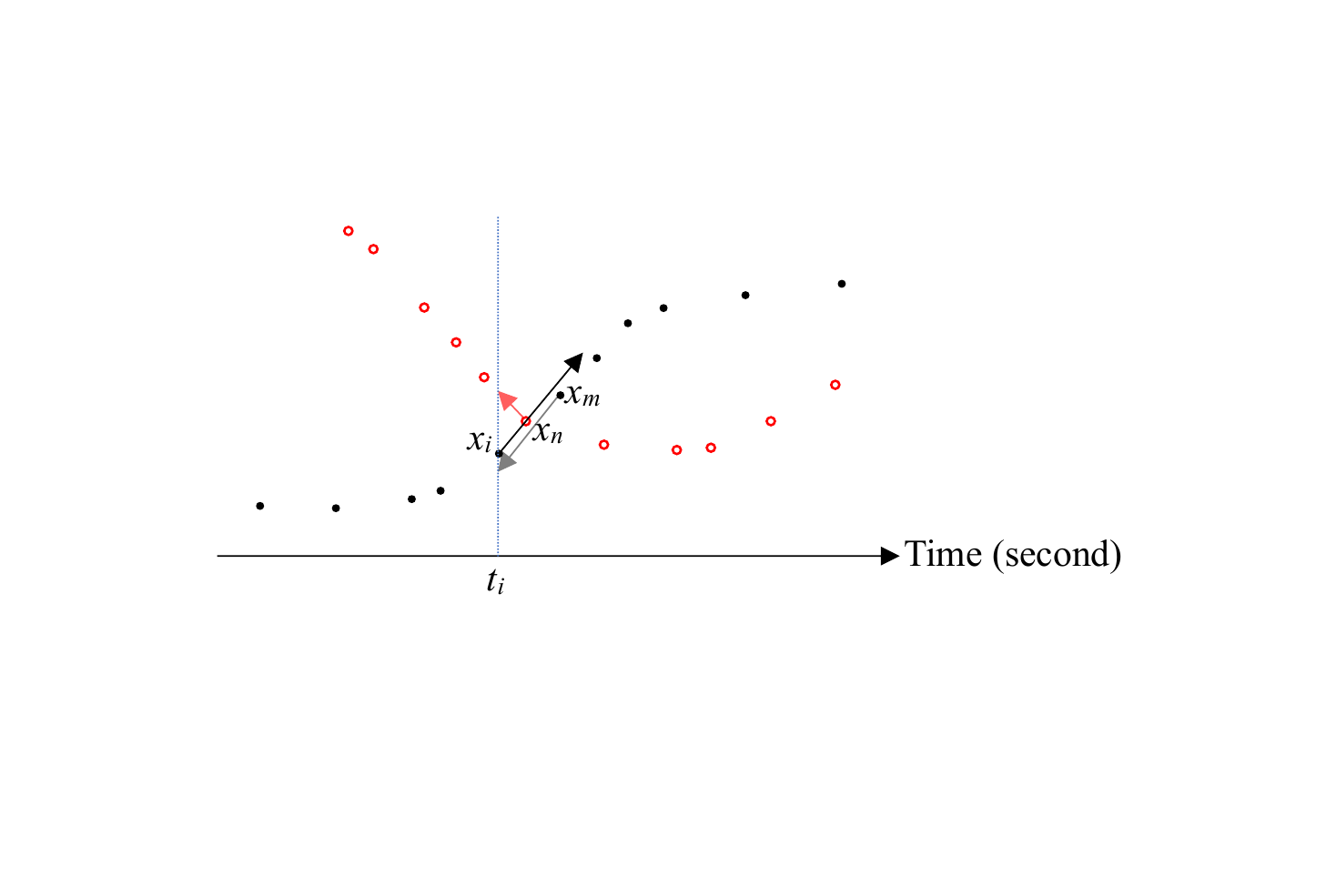}\vspace{-3cm}
    \caption{$d^+(x_i,x_n)$ is smaller than $d^+(x_n,x_m)$, so we need to involve $d^-$ to distinguish two crossing trajectories. One sees that $P^-(x_m)$ is close to $x_i$ while $P^-(x_n)$ is not, so one can use $d^-$ to eliminate $x_n$ to be the BPNP of $x_i$. This example explains why we define $d_{ij}:=\frac{1}{2}(d^++d^-)$.}
    \end{figure}

    {\it Case 2 for steady pairs}.
    
    Another type of the error distance is adapted to decide BPNP for pairs of steady points. In this case, since the space difference is tiny, we compute the distance with a smaller weight on time as follows.
    $$ d_0:=(2\cdot10^{-9}\cdot\mbox{time diff.})^2+\alpha^2(\mbox{latitude diff.})^2+(\mbox{longitude diff.})^2$$ to compute distance of $x_i$ and its candidates. Similar to the previous case, we kick off candidates with large angle $\theta_0$ w.r.t. the positive time direction $(1,0,0)$. Since the vessel is steady, we ask the angle to be smaller than $\arccos(0.95)\approx 18.19^\circ$. (See Theorem 3 in the next section for more explanations.)\\ 
    
    Remark: (i) Using $d^+$ and $d^-$ to do double checking significantly improves the accuracy at troublesome points. 
    
    (ii) In the definition of $d^+$, since 1 knot equals roughly $6\cdot10^{-6}$ degree in longitude per second and the vessel changes both its longitude and latitude, we use the factor $2\cdot10^{-6}$ to balance the time difference and space difference. For moving vessels, we use slightly lager factor $10^{-5}$ to compute the space-time angle, while for steady cases, it is better to choose small factor, such as $10^{-9}$ to prevent the domination of the time variable.  
    
    (iii) For steady vessels, we do not use $P^+(x_i)$ to find the BPNP of $x_i$, because the change of courses of floating vessels sometimes ruin the prediction position $P^+(x_i)$.\\

    \item Step 3. (Points whose BPNP is far away from the prediction position are endpoints of trajectories.) In step 2, we might find some points $y_k$'s with $d_{kj}=\infty$ or $d_0=\infty$ because they do not possess BPNP. These points are probably the true endpoints of trajectories of vessels. Besides these points, we choose a threshold number $N$ and select the first $N$ points $\{z_l\}_{l=1}^N$ which have largest normalized error distance $\tilde{d}_{lj}:=d_{lj}/(t_j-t_l)^2$ or $\tilde{d_{0}}:=d_{0}/(t_j-t_l)^2$. 
    Apparently these $N$ points very likely contain endpoints. If $z_{l'}:=BPNP(z_l)$ locates very near to $z_l$ in space, say $|z_{l'}-z_l|$ is less than $350$ meters and the (space-time) turning angle $$\varphi_{l}:=\frac{\overrightarrow{z_lz_{l'}}\cdot \overrightarrow{z_{l'}z_{l''}}}{\left|{\overrightarrow{z_lz_{l'}}}\right|\left|{\overrightarrow{z_{l'}z_{l''}}}\right|}, \mbox{ where } BPNP(z_{l'})=z_{l''},$$ is less than $\cos^{-1}(0.6)$, we treat $z_l$ as a turning point of a vessel and remove it from the bad point list $\{z_l\}_{l=1}^N$. The remaining bad points together with $y_k$'s are called abnormal points.\\ 
    
    Remark: In practice we found $N>30$ works well and our algorithm is very robust to this number.\\
    
    \item Step 4. Cluster all points by connecting each point with its BPNP, except for abnormal points. 
\end{itemize}

\section{Results}
We evaluated the results by the correct-neighbor rate that is defined as $
\sum_{i=1}^n I(Y_{i}=Y_j)/n,
$
where $Y_j$ is the label of the closest neighbor of $Y_i$.
For example, assuming that the predicted labels are $(1,2,1,2,1,2)$ and the left of each point is the closest neighbor,
then the correct-neighbor rate is $0$, but the overall label correctness rate is $50\%.\\$ 
\begin{table}
\centering
\caption{The correct-neighbor rates for each method in the three sets.}\label{tab1}
\scriptsize
\begin{tabular}{ |p{4cm}||p{2cm}|p{2cm}|p{2cm}|  }
 \hline
 Methods& Set 1 & Set 2 & Set 3\\
 \hline
 NPC Classification   &$0.9942$  &  $0.9881$   &$0.9942$\\
 NPC Clustering &   $0.9732$  &  $0.9481$   &$0.9842$\\
 CBTR ($N=50$) &$0.9985$ & $0.9981$&  $0.9971$\\
 LSTM &$0.6580$ & $0.6749$ &  $0.6534$\\
 EM clustering&$0.1580$ & $0.1749$ &  $0.1643$\\
 \hline
\end{tabular}
\end{table}

We compared the proposed clustering method with the EM algorithm \citep{Rubin1982}. The comparisons of their correct-neighbor rates and computational time are listed in Tables~\ref{tab1} and \ref{tab2}.
\begin{table}
\centering
\caption{The computational time for each method in the three sets in seconds.}\label{tab2}
\scriptsize
\begin{tabular}{ |p{4cm}||p{2cm}|p{2cm}|p{2cm}|  }
 \hline
 Methods& Set 1 & Set 2 & Set 3\\
 \hline
 NPC Classification   &$20$  &  $27$   &$23$\\
 NPC Clustering &   $15$  &  $16$   &$17$\\
 CBTR ($N=50$) &$20$ & $29$&  $18$\\
 LSTM    &278 & 405&  262\\
 EM clustering&   20  & 31&27\\
 \hline
\end{tabular}
\end{table} 
The time complexity of the proposed CBTR method is $O(nr)$ with the sample size $n$ and the neighborhood size $r$. Based on the design of the proposed clustering algorithm, we conclude the properties of the proposed CBTR method as follows.
\begin{lemma}
Given a vessel, the points are not connected with other vessels if and only if the changes of their features (longitude, latitude, time, speed, and direction) with the vessel are smaller than the changes between vessels.
\end{lemma}

\begin{theorem}\label{endpoint}
Using the proposed clustering-based trajectory reconstruction method, a point $x_i$ is determined to be an endpoint if one of the following situations occurs:

{\rm(i)} $x_i$ has no future points;

{\rm(ii)} the rescaled error distance of $x_i$ and its BPNP is larger than the threshold (which is determined by $N$), and either the turning angle  $\varphi_{ij}$ from $x_i$ to its BPNP is larger than $\cos^{-1}(0.6)\approx 53.13^\circ$ or the space distance of $x_i$ and its BPNP is larger than 350 meters.
\end{theorem}

From Theorem \ref{endpoint} above, we know that each $x_i$ is connected to its BPNP $x_j$ if it does have some future points and one of these future points, $x_j$, satisfies either  
(1) the error distance is less than the threshold or (2) the turning angle $\varphi_{i}\leq 53.13^\circ$ and $|x_i-x_j|\leq$ 350 meters. We remind the reader that $\varphi_{i}$ is an angle in space-time but not the angle on the earth.

Most points $x_i$ of a generic trajectory lie in the first category (1), because their BPNPs are usually the next point or the second next point, etc. Sometimes the vessel makes a turn somewhere with sparse record points, then the error distances between these points might be large. In this case, we use the second condition to restore the trajectory. When there are no other vessels nearby, this process works well. However, if there is another trajectory passing through the neighborhood, we have to prevent $x_i$ connecting to a BPNP which indeed belongs to this passing vessel. We observe that there are two different types of trajectory-passing and they should be treated separately as follows.

For a point $x_i$ and denote $x_j$ as one of its candidates for BPNP. We say that $(x_i,x_j)$ is a {\it moving pair} if the sum of speeds of vessels at $x_i$ and $x_j$ is less than $4$; Otherwise $(x_i,x_j)$ is called a {\it steady pair}. In the following theorems, all space-time vectors are defined as in Step 2.

\begin{theorem} (Similarity threshold for moving vessels)
For a point $x_i$, denote its predicting next position as $P^+(x_i)$ and its BPNP as $x_j$. If $(x_i,x_j)$ is a moving pair, then the space-time angle must satisfy $$\theta_{ij}=\frac{\overrightarrow{x_iP^+(x_i)}\cdot\overrightarrow{x_ix_j}}{\left|\overrightarrow{x_iP^+(x_i)}\right|\left|\overrightarrow{x_ix_j}\right|}\leq\cos^{-1}(0.1)\approx84.26^\circ.$$ This means every two points of moving vessels are not connected by CBTR if their space-time angle defined in step 2 is greater than $84.26^\circ$.
\end{theorem}

   \begin{figure}[H]
    \centering
    \includegraphics[scale = 0.4]{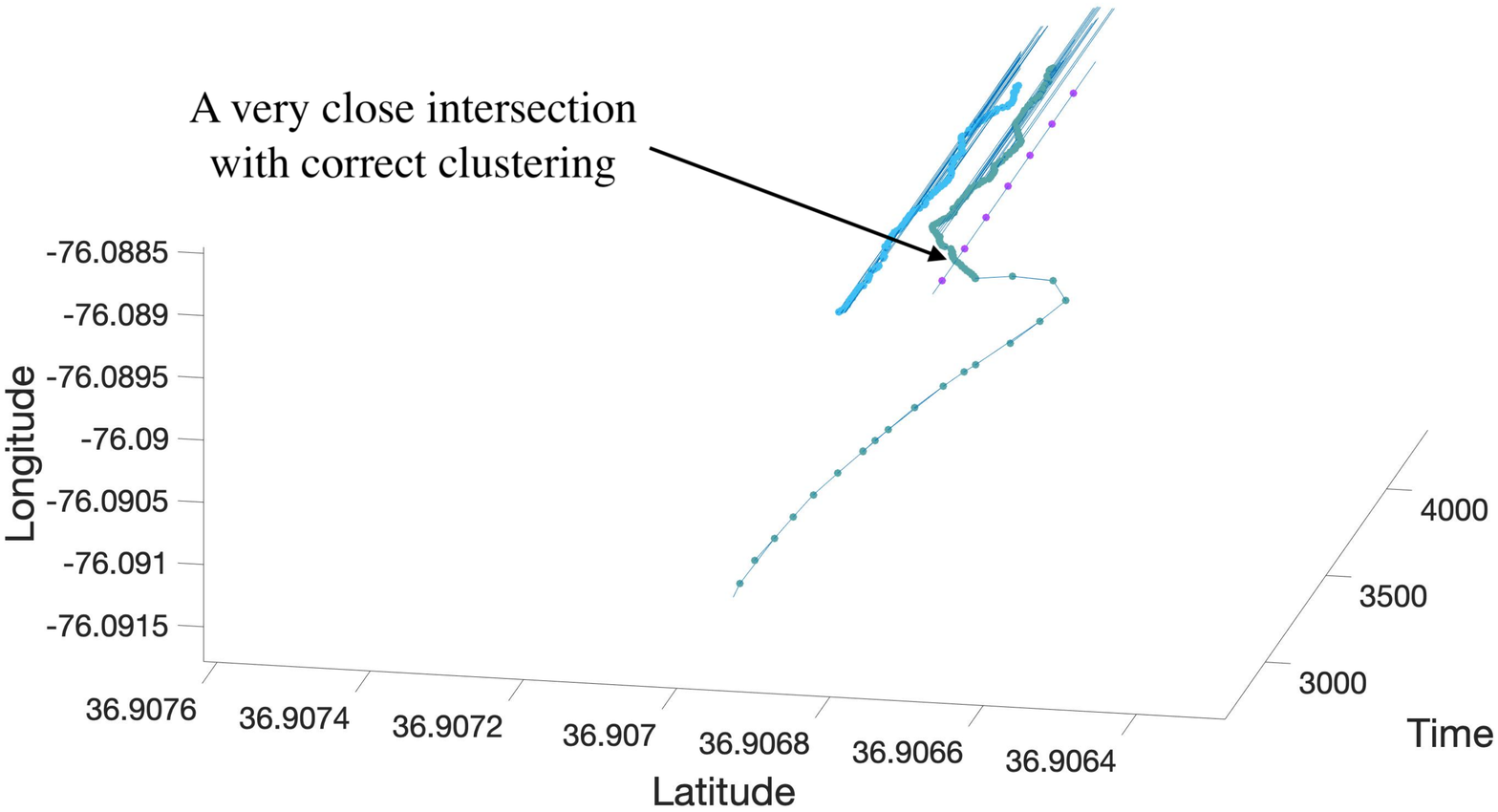}\vspace{0cm}
    \caption{Part of 3 trajectories are shown here, which are colored by VID and linked by CBTR. The vessel in green makes a dramatic turn and then docks between the vessel in purple and the vessel in light blue. We can avoid incorrect clustering at the intersection of vessels because we use double checking distance $d_{ij}$ and the angle condition in Theorem 2.}
    \end{figure}

On the other hand, we have the following theorem to prevent merging two steady close vessels.

\begin{theorem} (Similarity threshold for steady vessels)
For a point $x_i$, denote its BPNP as $x_j$. If $(x_i,x_j)$ is a steady pair, then the space-time angle between the time direction $\overrightarrow{\bf u}=(1,0,0)$ and $\overrightarrow{x_ix_j}$ must satisfy $$\theta_0:=\frac{\overrightarrow{\bf u}\cdot\overrightarrow{x_ix_j}}{\left|\overrightarrow{\bf u}\right|\left|\overrightarrow{x_ix_j}\right|}\leq\cos^{-1}(0.95)\approx18.19^\circ.$$ This means that, if we consider 1000 seconds in Step 1, then two steady vessels are not merged by CBTR whenever they park apart from each other more than 1.14 kilometers. (One can replace 0.95 by 0.9995 and the 1.14 kilometers becomes 11.7 meters. Since a steady boat might float around in 11 meters as we observed from the data, it is in vain to increase the accuracy further.) 
\end{theorem}

    
    \begin{figure}[H]
    \centering\vspace{-1cm}
    \includegraphics[scale = 0.4]{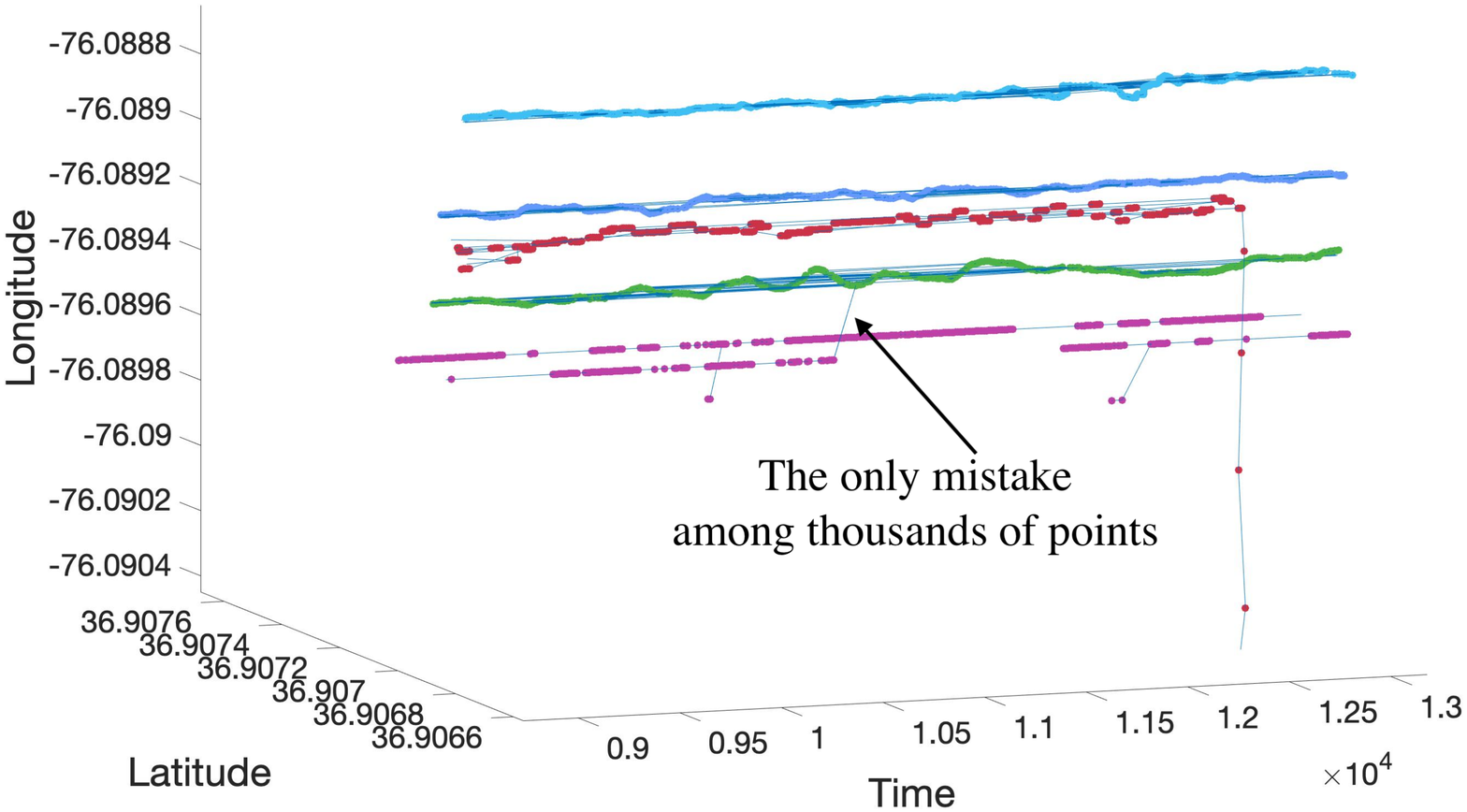}\vspace{-0.5cm}
    \caption{CBTR is highly accurate and can distinguishes vessels parking nearby. Note that 0.0001 degree of longitude is about 8 meters.}\vspace{-4cm}
    \end{figure}
    
 \begin{figure}[H]
    \centering
    \includegraphics[scale = 0.55]{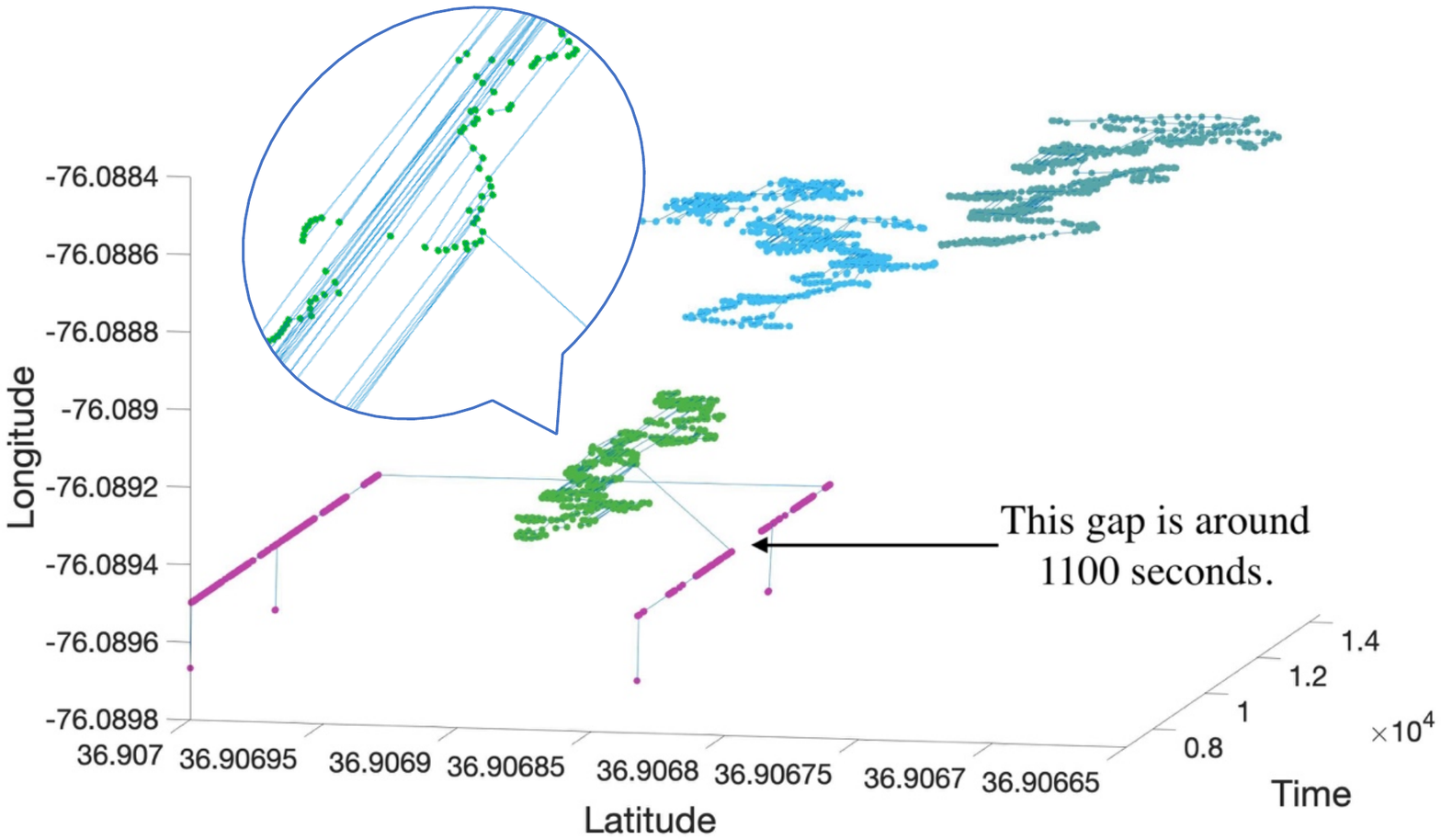}\vspace{-4cm}
    \caption{The only miss-clustered points in Fig. 5 can be seen from another perspective. Since the AIS data of the purple vessel in purple is irregular and may be a machine error (it has many disconnected points), thus it could not be clustered correctly. Note that, in step 1, we collect candidates for BPNP up to 1000 seconds and the gap in the purple trajectory is more than 1000.}
    \label{fig6}
    \end{figure}
Figure~\ref{fig6} illustrates the case that CBTR fails.
When the purple vessel that is a steady boat drifting randomly has adjacent points with the time gap larger than the search range (e.g. 1000 seconds gap), the step 2 of CBTR cannot connect them. Nevertheless, if the nearby green vessel has points within the search range of the purple vessel, then the green point which is nearest (measured by $d_0$ in step 2) to the purple break point will be connected to the break point by CBTR. It leads to incorrect merging with other vessels as we observed in the experiments.

At last, we demonstrate our result by the clustering plots Fig. 7, 8, and 9, accompanied with two numbers: jumps and merges. The former shows how many times CBTR breaks trajectories wrongly and the later shows how many incorrect links it makes. The sum of jumps and merges is a good index to judge the performance of CBTR. Moreover, we have the following formula:
$$\mbox{Actual number of vessels } =  \mbox{ number of predicted clusters } + \mbox{ merges } - \mbox{ jumps}.$$

\section{LSTM Method}
Long Short-Term Memory \citep{hochreiter1997long} is a type of Recurrent Neural Networks (RNNs). LSTMs are an example of a recurrent neural network. A recurrent neural network is a neural network that has feedback loops; that is, a neural network that introduces cycles which allows time-dependent problems to be solved. Technically, we mean that the outputs (i.e. previous outputs) can be used as an input to help model the current output. More generally, problems that have a fundamental order can be solved. One thing to keep in mind here is that LSTMs are capable of modeling sequences of different lengths, and this is ideal as vessel paths often have a different number of points. The output for an LSTM at time $t$ can be denoted by $\mathbf{h}_t = f_W(\mathbf{h}_{t-1},\mathbf{x}_t)$ where $f_W(\cdot)$ is some pre-defined activation function like $\tanh$ or $ReLU$. This allows the LSTM to model linear or nonlinear relationships over time. The key advantage in using an LSTM lies in how the model is updated. Specifically, there are gating units in each memory cell. A forget gate is given by $0\leq \sigma(W_f*[h_{t-1},x_t] + b_f) \leq 1$, and the value determines the extent to which previous information is kept or forgotten, hence the name. Values closer to 1 mean that much of the information is kept whereas values closer to 0 mean that much of information is discarded. Notice here that when the weights $W_f$ are larger, then most of the information is kept whereas as when the weights decay, then the forget gate takes a smaller value and thus the information is discarded. The input gate determines which entries in the cell state should be updated. The previous cell state $C_{t-1}$ is multiplied (i.e. Hadamard or component-wise) by the forget gate output and then added to updated cell state $C_t$ multiplied by the new input information. This takes the following form: $C_t = f_t*C_{t-1} + i_t*\tilde C_t$. Finally, the output gate uses a sigmoid function of the previous state and current information: $o_t = \sigma(W_0[h_{t-1},x_t] + b_0)$ and $h_t = o_t*\tanh(C_t)$. To summarize, LSTMs adapt by learning crucial patterns while forgetting unnecessary information through a series of filters and transformations.
\subsection{LSTM Next Point Prediction}
LSTMs are convenient for the AIS prediction problem as they can naturally be adapted to multi-target learning and are capable of learning both simple and complex patterns. Here we can think of the timestamp, latitude, longitude, speed, and direction, all at time $t$, as response variables whereas the predictor variables (i.e. inputs to the LSTM) are the timestamp, latitude, longitude, speed, and direction at time $t-1, t-2, \cdots, t-k$. We train an LSTM using lagged versions of the timestamp, latitude, longitude, speed, and direction (i.e. time $t-1, t-2.\cdots,t-k$) in order to predict the timestamp, latitude, longitude, speed, and direction at one time point in the future (i.e. time $t$). The goal here is to attempt to predict all characteristics of a vessel automatically using previous information. The architecture and tuning were accomplished via trial an error using a random 20\% validation sample. The characteristics of the LSTM  are the following: an input dimension of 5 (i.e. timestamp, latitude, longitude, speed, and direction are lagged by $k=1$ time unit), $1$ hidden layer, $250$ hidden units using the Rectified Linear Unit (ReLU) activation function: $\max(0,x)$, and $5$ output nodes (i.e. timestamp, latitude, longitude, speed, and direction at time $t$). Additional values for the number of lags were tried, but the performance was essentially unchanged and different activation functions were tried and tended to produce inferior results. The software used was the keras library in Python \citep{Charles2013}.

The predicted path is derived from using the LSTM prediction at the next time step. Formally, the algorithm is the following. \begin{itemize}
    \item Step 1: Train an LSTM model $LSTM(\mathbf{X})$ where $\mathbf{X}$ is a matrix of lagged predictor variables
    \item Step 2: Path Initialization: from the nodes not selected in a path, pick the node with the oldest time $\mathbf{x}_{t-1}$
    \item Step 3: Predict the next node $\widehat{\mathbf{x}}_t = LSTM(\mathbf{x}_{t-1})$
    \item Step 4: Find the nearest neighbor to $\widehat{\mathbf{x}}_t$ within some time interval $(time_{t-1} - q_L, time_{t-1} + q_U)$ where $time_{t-1}$ is the time for the observed node $\mathbf{x}_{t-1}$.
    \item Step 5: Add the nearest neighbor to the predicted path
    \item Repeat Steps 3-5 until the (shifting) time window is empty
    \item Go back to Step 2. Repeat until every point is assigned to a cluster. 
\end{itemize}
The results from the LSTM using all five variables as outputs seem to indicate that this approach is unable to distinguish the different boat paths.

\begin{figure}[H]\label{01all}
    \centering
    \includegraphics[scale = 0.4]{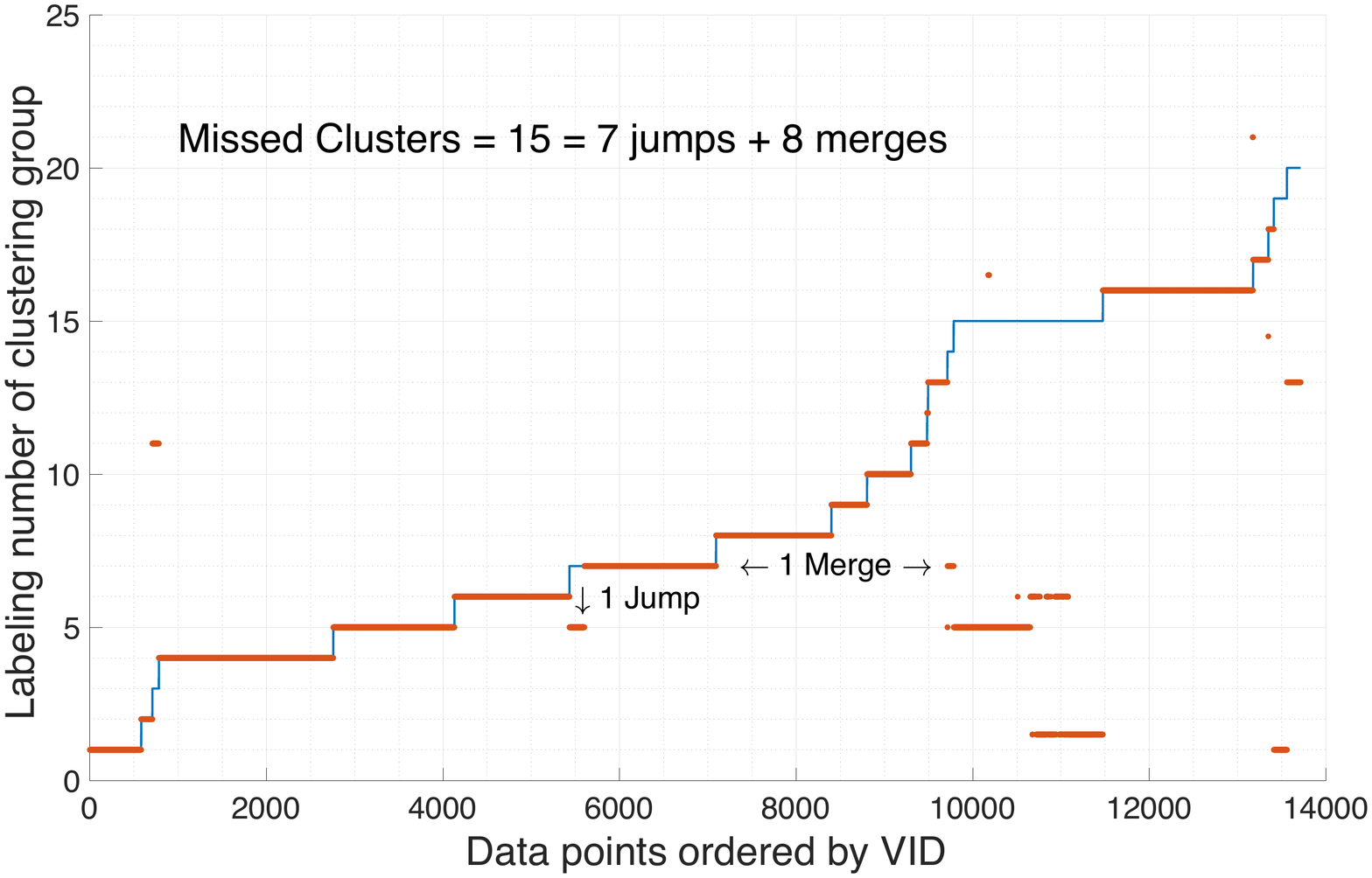}
    \caption{The AIS data set 1: There are 20 vessels that are clustered into 19 groups via threshold $N=50$ using CBTR with a $1,000$ second threshold in Step 1. The actual VIDs are presented by horizontal blue segments whereas the red segments indicate CBTR-clusters. 
    The $y$-axis value represents the predicted label of the clusters. The correctness rate is $99.71\%$, which means that most of the red points are correctly connected into segments. The ``Missed Clusters" are the clusters that are either separated from the true clusters or connected with other clusters. For instance, vessel no.3 is classified into the same cluster to vessel no.11. This is called a {\it merge} but not counted as a {\it jump} because vessel no.3 does not separated into multiple groups. On the other hand, some portion of vessel no.7 is connected incorrectly to vessel no.5. This contributes 1 jump and 1 merge. One can see that there are 5 groups, thus 4 jumps, in vessel no.15. In fact, the record of vessel no.15 is problematic and thus the classification is bad.}
\end{figure}

\begin{figure}[H]
    \centering
    \includegraphics[scale = 0.4]{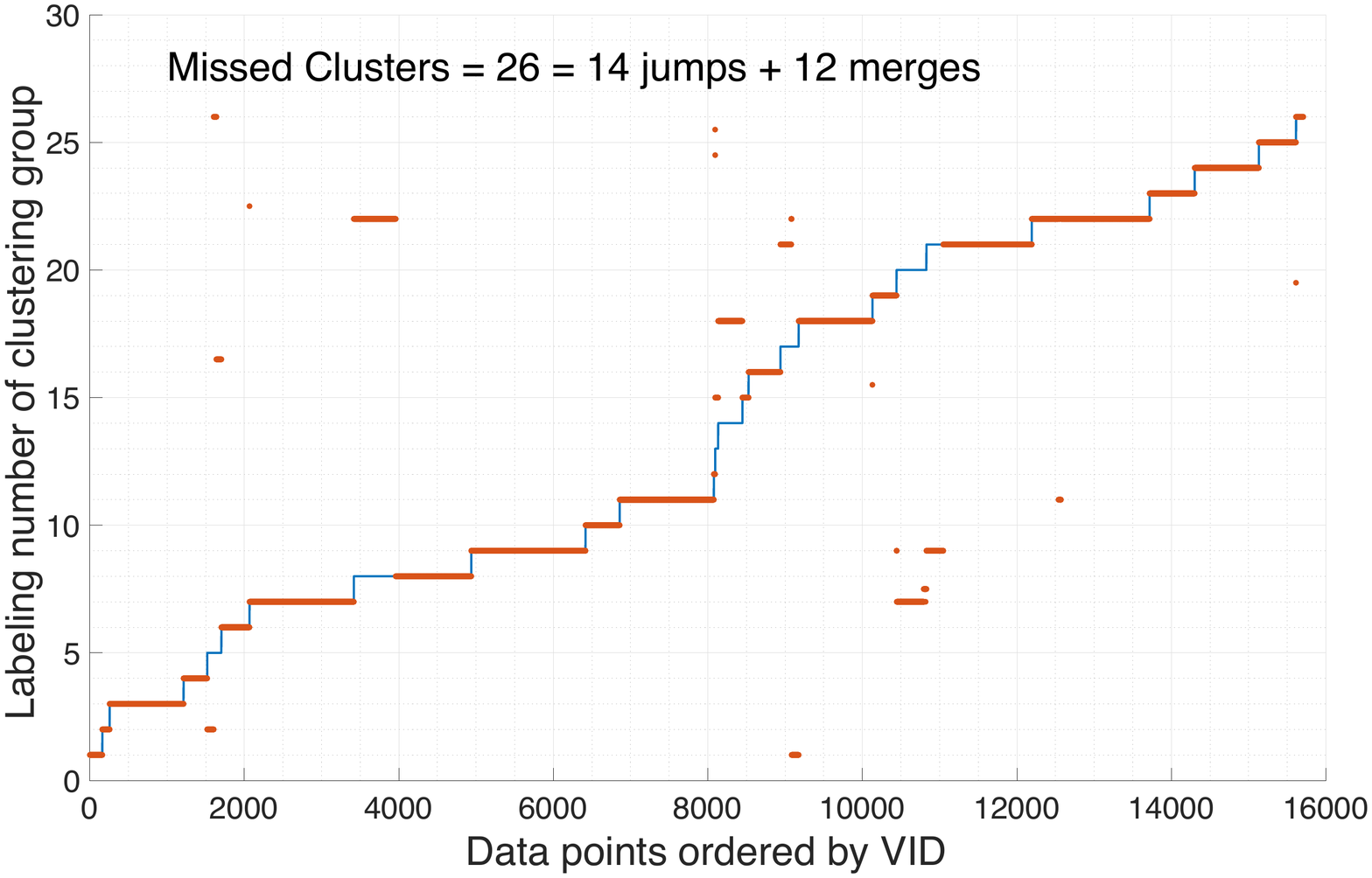}
    \caption{The AIS data set 2: There are 26 vessels that are clustered into 28 groups via threshold $N=50$ using CBTR with a $1,000$ second threshold. The correctness rate is $99.85\%$.}

    \includegraphics[scale = 0.4]{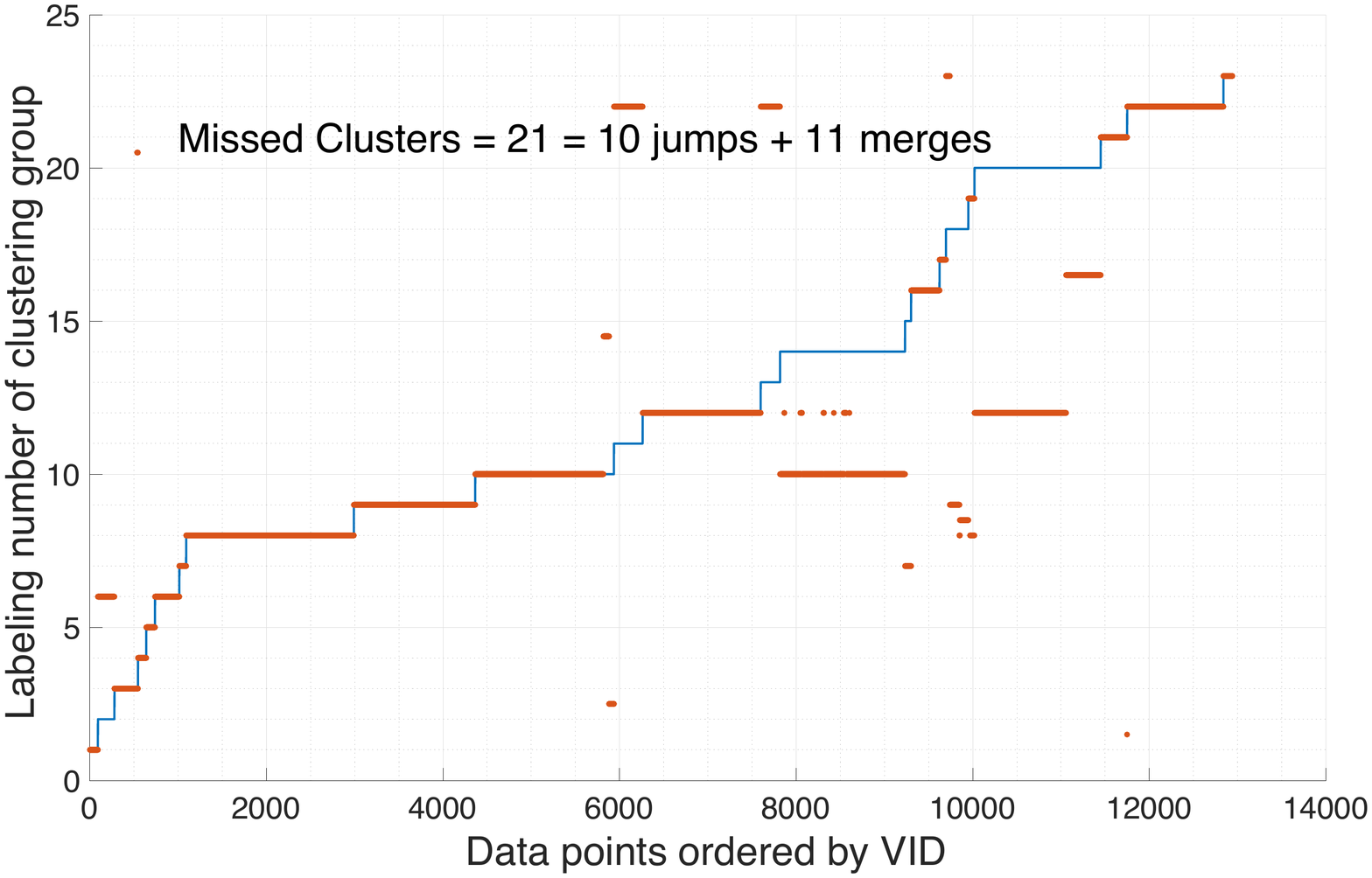}
    \caption{The AIS data set 3: There are 23 vessels that are clustered into 22 groups via threshold $N=50$ using CBTR with a $1,000$ second threshold. The correctness rate is $99.81\%$.}
\end{figure}

\begin{figure}[H]
    \centering
    \includegraphics[angle=-90, scale = 0.75]{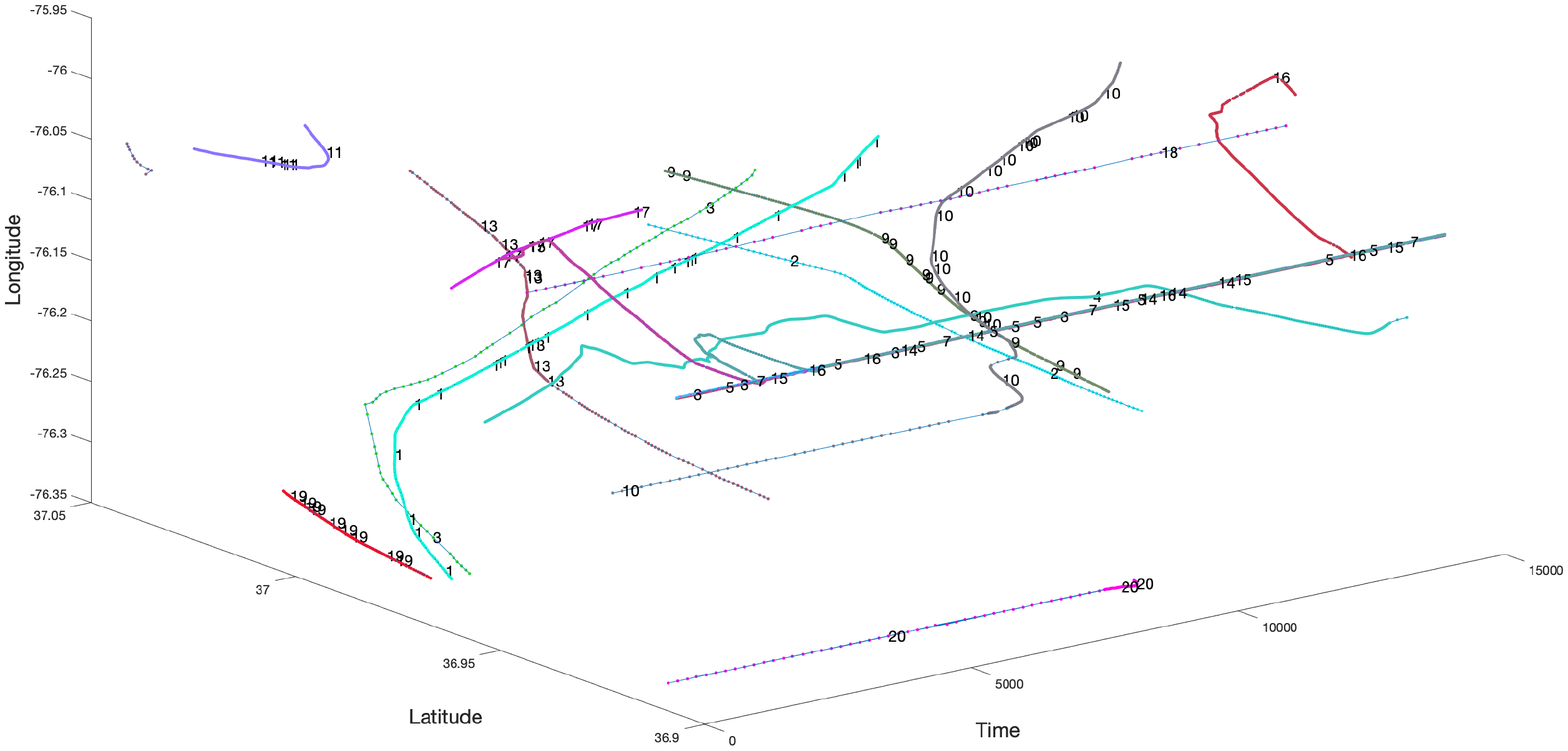}
    \caption{Each vessel trajectory in data set 1 is plotted with a unique color for each VID, and the trajectory reconstruction was accomplished via CBTR.}
\end{figure}

\begin{figure}[H]
    \centering
    \includegraphics[angle=-90, scale = 0.75]{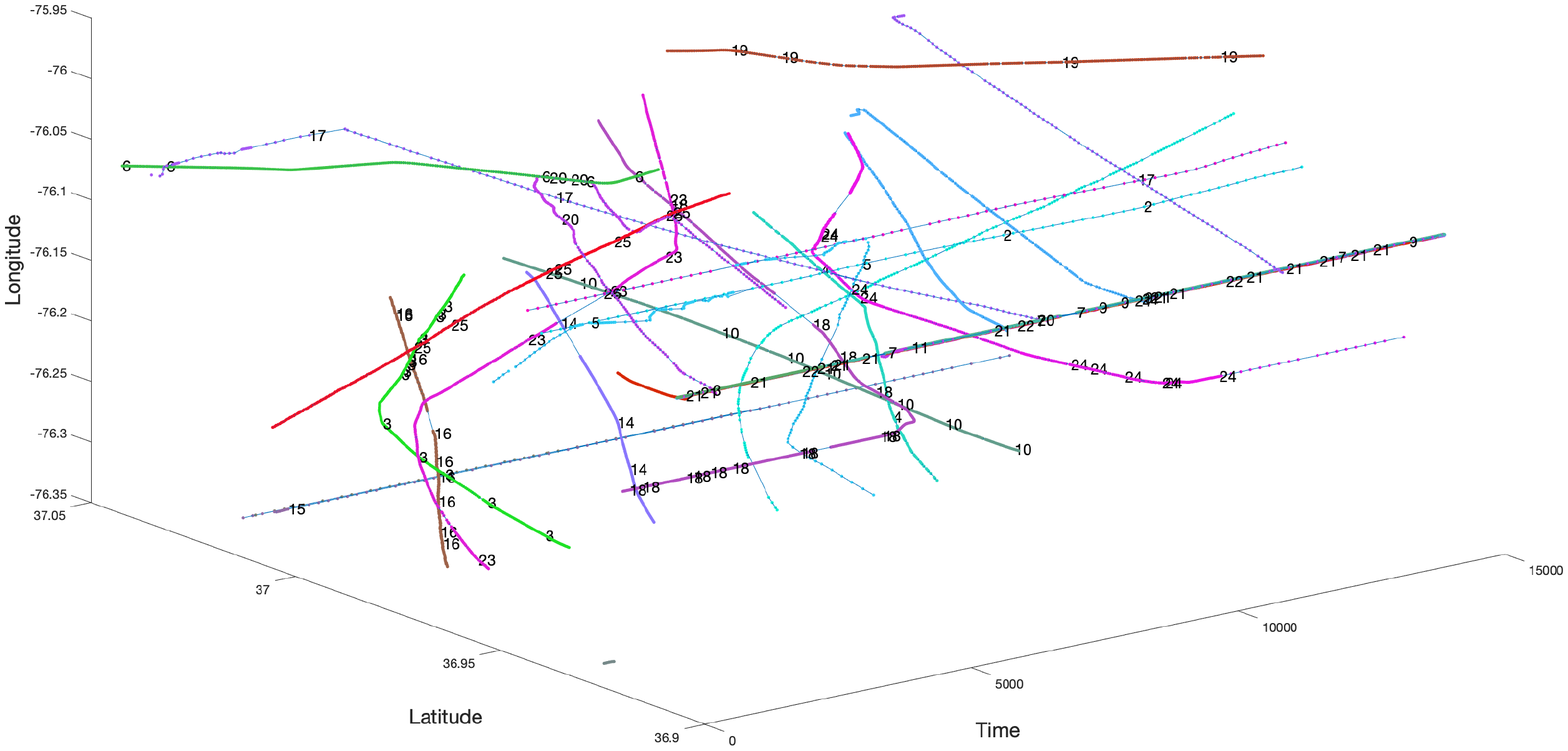}
    \caption{Each vessel trajectory in data set 2 is plotted with a unique color for each VID, and the trajectory reconstruction was accomplished via CBTR.}
\end{figure}

\begin{figure}[H]
    \centering
    \includegraphics[angle=-90, scale = 0.75]{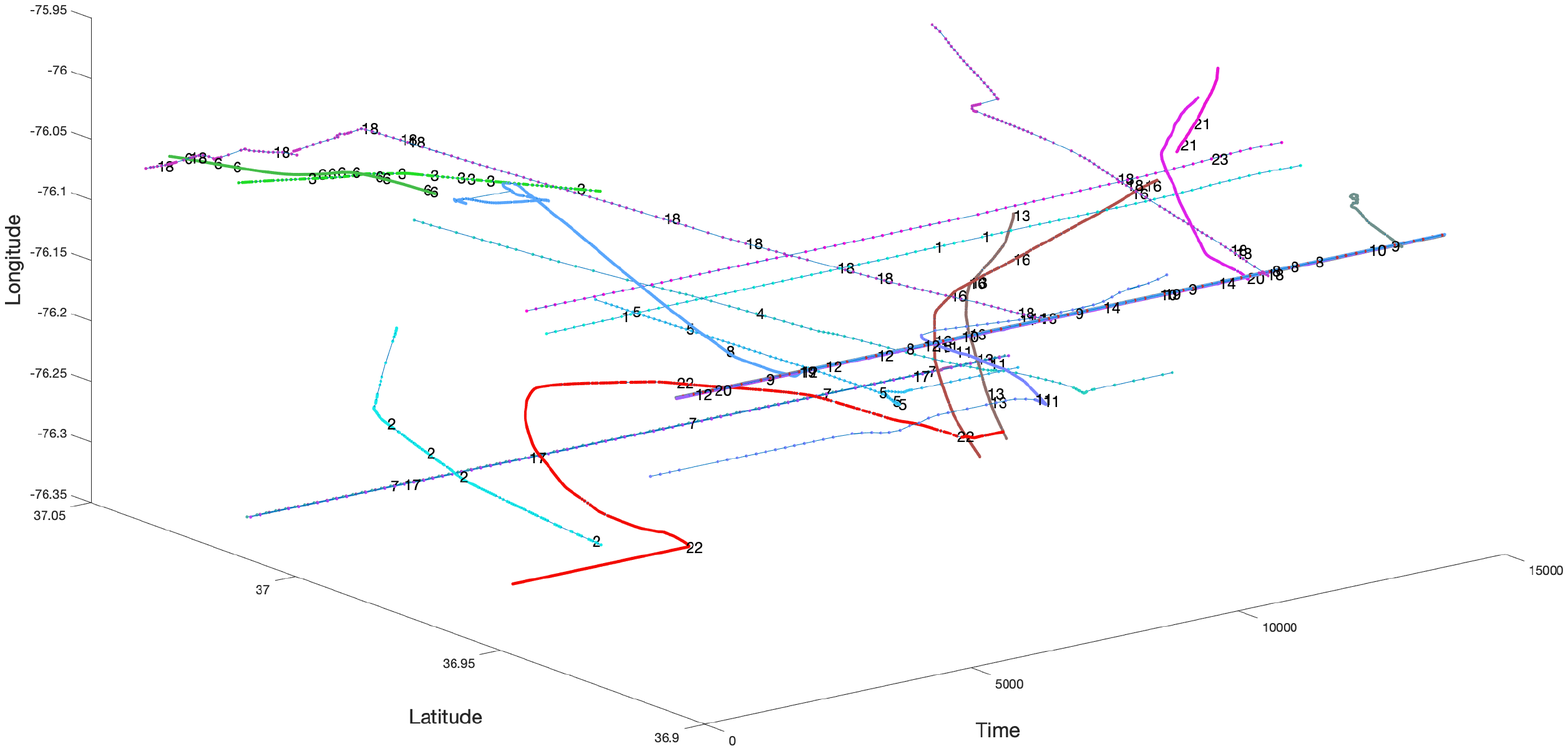}
    \caption{Each vessel trajectory in data set 3 is plotted with a unique color for each VID, and the trajectory reconstruction was accomplished via CBTR.}
\end{figure}

\section{Conclusion and Discussion}
The proposed method vitally improves the reliability and accuracy of prediction of vessel trajectories within a specified time.
This article presents a real-time algorithm of ship movement trajectory prediction which utilizes the local information of the ship’s positions. In addition, the algorithm does not require a training model. It provides a fast, reliable, and accurate trajectory prediction which is desirable in the navigational decision support system.

\subsection{Discussion on the performance of CBTR}

Figure 12 is a high resolution zoom-in picture of the AIS data set 1. One can see three vessels, no. 5, 6, and 7, stay still and close to each other at the beginning. Then vessel no. 7 leaves at time 1500, encounters no. 15 somewhere and make a drastic turn, where the algorithm misses a single track. Vessel no. 15 has a vibrating trajectory around and seems not be recorded correctly. The vessel comes and parks near vessels no. 5 and 6. An incorrect link occurs when it approaches no. 6. At the endpoint of this trajectory, we see that it connects with no. 5 incorrectly. This happens when the incorrect link meets the trajectory of no. 5 in a small angle. Since all these happens in a extremely tiny space region and the data of vessel no. 15 is noisy, it is difficult to set them apart.

\begin{figure}[H]\label{01local}
    \centering
    \vspace{-1cm}
    \includegraphics[scale = 0.4]{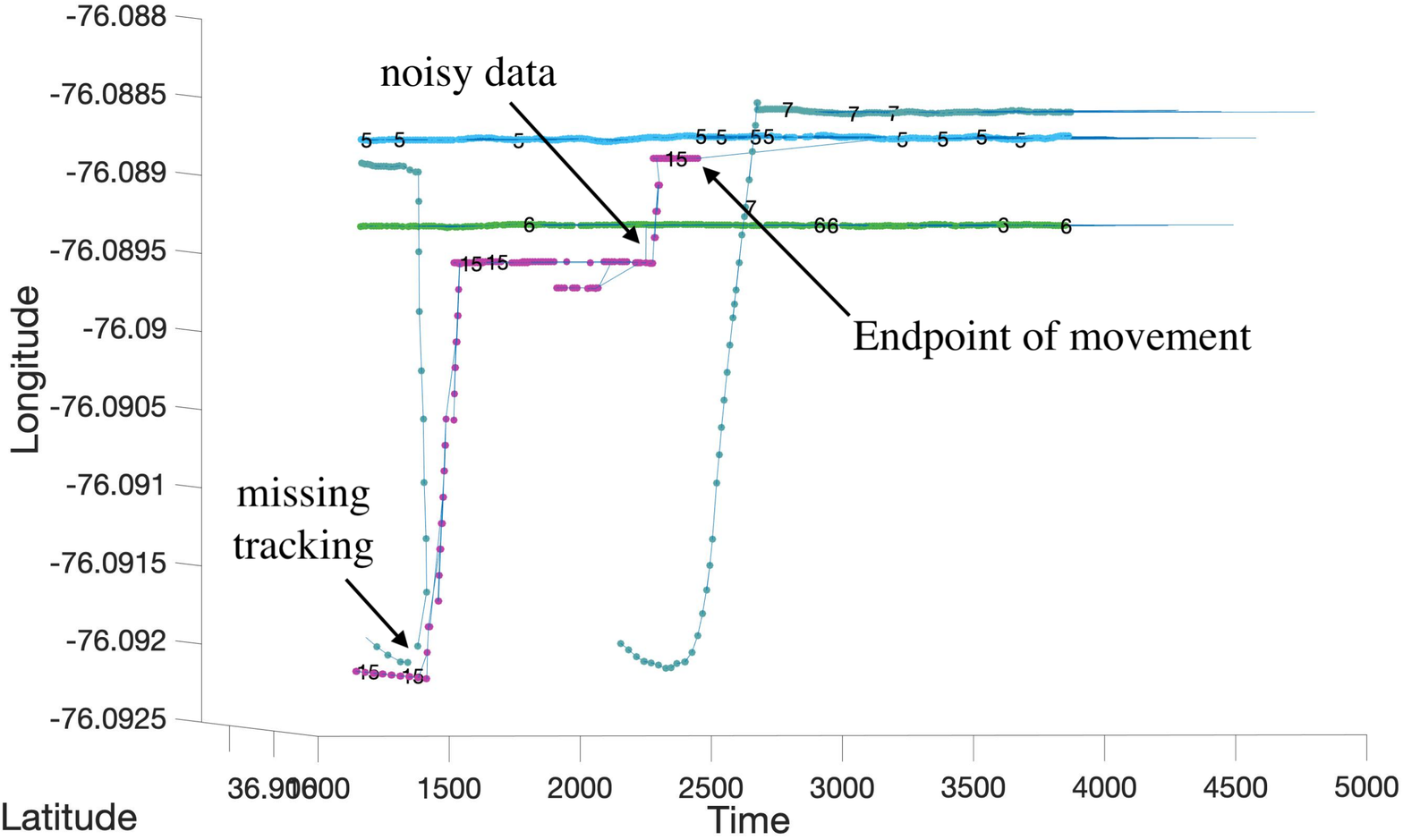}
    \caption{A zoom-in part of the trajectory reconstruction using the CBTR in the AIS data set 1.  The incorrect reconstruction is due to instant drastic direction changes (no.7 is broken at time 1578), end of the trajectory (no.15 connects to no.5), or noisy data (a point of no.15 at time 2391 connects to no.6).}
\end{figure}

CBTR performs very well except for highly noisy data and endpoints of trajectories. As Fig.13 suggests, if we change the time threshold from 1000 to 300, the incorrect link between no. 15 and no. 5 does disappear. In this case, the missed clusters is 37 (26 jumps + 11 merges) and the algorithm spends 9 seconds. The number of jumps becomes lager because some gaps of successive points are far apart than 300 seconds. For instance, there are 26 such gaps in Data set 1, half can be observed on the trajectory of vessel no. 18 in Fig. 4. Once we break these links, these points either be determined as end points or would find false replacements to be their next points.

A detailed analysis shows that almost all errors in the result by CBTR are of three types: missing tracking of moving vessels, wrongly connected steady vessels, and large gap in time (i.e., $>1000$). We conclude that, for a moving vessel, CBTR rarely makes incorrect link and merges the vessel into another one. CBTR works accurately except for the suddenly dramatic trajectory change.
On the other hand, CBTR never splits a steady vessel into multiple clusters in our experiments.

\subsection{Discussion on the performance of the LSTM Path Prediction }
The performance of the LSTM next point prediction method is fundamentally dependent on the LSTM’s ability to predict the properties of the node at the next time point. That is, it must be able to accurately predict the timestamp, latitude, longitude, speed, and direction at some future point in time. The nearest neighbor search between the predicted node and the observed nodes only occurs within a pre-defined time window, but the number of potential nodes that can be selected in this window is large enough that mistakes can and will be made. An inspection of the LSTM predictions and the resulting nearest neighbor search indicate that much of the error is related primarily to two factors: some vessels rapidly change their speed and direction while simultaneously other vessels that were previously similar to the rapidly changing vessel do not change their speed or direction suddenly and this results in misclassifications. An example of this is vessel no. 8 and no. 5 in the first dataset. The second source of error seems to be that the LSTM predictions are often not precise enough, and in combination with a larger number of candidates within each time window (i.e. the time window in the nearest neighbor search), mistakes are made. Another limitation is the relatively small amount of data. LSTM models are known to require a large amount of data in order to be effective, so the relatively small size of the individual AIS training datasets also is a contributing factor to the LSTMs performance.       

\subsection{Experiments by sampling}
We conducted experiments to evaluate the robustness of the proposed clustering method by (1) deleting every fifth point of each five points (i.e. removing the fifth, tenth, ..., etc.) and (2) deleting every second point of each two points (i.e. removing the second, fourth, ..., etc.). The experiments remove $20\%$ and $50\%$ points respectively. 
For the AIS data set 1, 2, 3 down-sampled by method (1), the correct-neighbor rates are $0.9960$, $0.9951$, and $0.9942$, respectively.
$0.9924$.
For the AIS data set 1, 2, 3 down-sampled by method (1), the correct-neighbor rates are For the AIS data set 1, 2, 3 down-sampled by method (1), the correct-neighbor rates are $0.9924$, $0.9910$, and $0.9892$, respectively. The removed points cause larger gaps between adjacent points in a vessel's path. Therefore, the more the points are removed, the lower the correct-neighbor rates are.
The number of jump can be reduced if we take an upper bound lager than 1000 in Step 1, which means that we consider more candidates when select BPNP. However, this would increase the number of merges at the same time.

The proposed CBTR method successfully reconstructs trajectories points without using a training set. 
Step 2 of the proposed CBTR is the spirit of our method, since it
uses the predicted forward and backward positions to measure the differences between two adjacent points. This method evaluate goodness of fitted path (projected positions) instead of using the static point information (location, time, speed, angle).
In step 2 of the CBTR algorithm, $x_i$ and $x_j$ within a reasonable time neighborhood (e.g. $t_i+1\leq t_j\leq t_i+1000$ seconds) are connected sequentially by minimizing the proposed error distance through the predicted next position $P^+(x_i)$ and the predicted previous position $P^-(x_j)$ instead of measuring the distance between $x_i$ and $x_j$. This step measures the goodness of fit of the predicted positions which are locally fitted positions using the location, time, speed, and angle of the current points $x_i$ and $x_j$. Apparently, if $x_i$ and $x_j$ belong to the same vessel, the corresponding $P^+(x_i)$ and $P^-(x_j)$ should be closet to each other. Therefore, this method is suitable for applying to moving-point data lacking for well-labelled vessels or containing new joint vessels or only few points of a vessel with small spatial and temporal gaps. When the spatial and temporal gaps within moving points of a vessel increase, the discrepancies within each vessel increase as well. 
 
We quantify the sufficient and necessary conditions that the proposed CBTR algorithm clusters the vessel paths correctly in the following theorem.
\begin{theorem}
Using the CBTR algorithm, a point $x_i$ is connected to its actual successive point $x_j$ if and only if either
\\
{\rm (A)} $x_i$ and $x_j$ form a moving pair; Conditions (i), (ii-1), (iii-1) hold; Either $\tilde d_{ij}$ is less than the threshold, or (iv), (v) hold; 

or
\\
{\rm(B)} $x_i$ and $x_j$ form a steady pair; Conditions (i), (ii-2), (iii-2) hold; Either $\tilde d_{0}$ is less than the threshold, or (iv), (v) hold.

Conditions are listed as follows:
\\
{\rm(i)} the time difference between adjacent points $x_i$ and $x_j$ satisfies $0<t(x_j)-t(x_i)\leq 1000$.
\\
{\rm(ii-1)} the rescaled error distance $\tilde d_{ij}:=\frac{1}{2}(d^++d^-)/|t_j-t_i|^2$ emanated from $x_i$ achieves its minimum at $x_j$.
\\
{\rm(ii-2)} the rescaled error distance $d_0$ emanated from $x_i$ achieves its minimum at $x_j$ and the value is less than the threshold.
\\
{\rm(iii-1)}  the space-time angle between $\overrightarrow{x_iP(x_i)}$ and $\overrightarrow{x_ix_j}$ must satisfy $$\theta_{ij}:=\frac{\overrightarrow{x_iP(x_i)}\cdot\overrightarrow{x_ix_j}}{\left|\overrightarrow{x_iP(x_i)}\right|\left|\overrightarrow{x_ix_j}\right|}\leq\cos^{-1}(0.1)\approx84.26^\circ.$$ 
\\
{\rm(iii-2)} the space-time angle between the time direction $\overrightarrow{\bf u}=(1,0,0)$ and $\overrightarrow{x_ix_j}$ must satisfy $$\theta_0:=\frac{\overrightarrow{\bf u}\cdot\overrightarrow{x_ix_j}}{\left|\overrightarrow{\bf u}\right|\left|\overrightarrow{x_ix_j}\right|}\leq\cos^{-1}(0.95)\approx18.19^\circ.$$ 
\\
{\rm(iv)} the (space-time) turning angle $\varphi_{i}$ is less than $\cos^{-1}(0.6)\approx 53.13^\circ$.
\\
{\rm(v)} the space distance of $x_i$ and $x_j$ is less than 350 meters.
\label{thm4}
\end{theorem}
Theorem~\ref{thm4} can be extended and applied to general cases of events and time for moving points represented in a three-dimensional spaces of the time difference between points, the rescaled error distance, and the space-time angle which narrows down the search range for the next point with candidates which fit the predicted path well, and hence improves the clustering accuracy.    
\section*{Acknowledgements}
This research was supported in part by the National Science Foundation, Grant Award No. 1924792.

\newpage
\bibliographystyle{spbasic_updated}      
\bibliography{sample_library}   

\end{document}